\documentclass[conference]{IEEEtran}
\IEEEoverridecommandlockouts
\usepackage{cite}
\usepackage{amsmath,amssymb,amsfonts}
\usepackage{algorithmic}
\usepackage{graphicx}
\usepackage{textcomp}
\usepackage{xcolor}
\usepackage{multirow}
\usepackage{todonotes}
\usepackage{amsmath} 
\usepackage{hyperref}

\usepackage{xcolor}

\def\BibTeX{{\rm B\kern-.05em{\sc i\kern-.025em b}\kern-.08em
    T\kern-.1667em\lower.7ex\hbox{E}\kern-.125emX}}
\begin{document}

\title{Functional Knowledge Transfer with Self-supervised Representation Learning\\
}

\author{
    \IEEEauthorblockN{
        Prakash Chandra Chhipa\textsuperscript{1,*},
        Muskaan Chopra\textsuperscript{2,a},
        Gopal Mengi\textsuperscript{2,b},
        Varun Gupta\textsuperscript{2},
        Richa Upadhyay\textsuperscript{1},\\
        Meenakshi Subhash Chippa\textsuperscript{1,c},
        Kanjar De\textsuperscript{1},
        Rajkumar Saini\textsuperscript{1},
        Seiichi Uchida\textsuperscript{3} and
        Marcus Liwicki\textsuperscript{1}\\
        \textit{\textsuperscript{1} Lule\aa~Tekniska Universitet, Lule\r{a}, Sweden}\\
        \textit{\{prakash.chandra.chhipa, richa.upadhyay, kanjar.de, rajkumar.saini, marcus.liwicki\}@ltu.se}\\ \textsuperscript{c} meechi-2@student.ltu.se\\
        \textit{\textsuperscript{2} CCET, Punjab University, Chandigarh, India}\\
        \textit{\{\textsuperscript{a} co19342, \textsuperscript{b} co20320, varungupta\}@ccet.ac.in}\\
        \textit{\textsuperscript{3}Human Interface Laboratory, Kyushu University, Fukuoka, Japan}\\
        \textit{uchida@ait.kyushu-u.ac.jp}\\
        \textit{\textsuperscript{*}Corresponding author - prakash.chandra.chhipa@ltu.se}
     }
    }

\maketitle

\begin{abstract}
This work investigates the unexplored usability of self-supervised representation learning in the direction of functional knowledge transfer. In this work, functional knowledge transfer is achieved by joint optimization of self-supervised learning pseudo task and supervised learning task, improving supervised learning task performance. Recent progress in self-supervised learning uses a large volume of data, which becomes a constraint for its applications on small-scale datasets. This work shares a simple yet effective joint training framework that reinforces human-supervised task learning by learning self-supervised representations just-in-time and vice versa. Experiments on three public datasets from different visual domains, Intel Image, CIFAR, and APTOS, reveal a consistent track of performance improvements on classification tasks during joint optimization. Qualitative analysis also supports the robustness of learnt representations. Source code and trained models are available on GitHub \footnote{\href{https://github.com/prakashchhipa/Functional_Knowledge_Transfer_SSL}{https://github.com/prakashchhipa/Functional\_Knowledge\_Transfer\_SSL}}. 
\end{abstract}


\section{Introduction}
The concept of functional knowledge transfer \cite{inductive} has been explored for multi-task learning problems in computer vision~\cite{mtl98, mtl_bene, mtl_survey} in the context of simultaneous training and joint optimization of multiple tasks. Typically functional knowledge transfer is employed for end-to-end joint training and optimization of multiple supervised learning tasks. 
Representational knowledge transfer, where pretraining and downstream task learning is done sequentially, has been thoroughly investigated and shown success in self-supervised learning.
So far, functional knowledge transfer in self-supervised learning has not been studied, leaving a research gap.

\begin{figure}[t] 
    \includegraphics[width=.9\columnwidth]{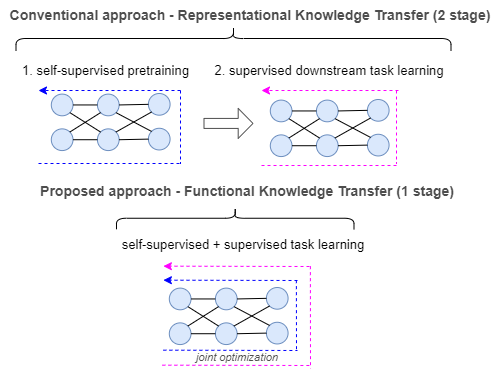}
    \centering
    \caption{Figure compares the proposed functional knowledge transfer approach in context of self-supervised learning and supervised task learning with conventional representational knowledge transfer approach where self-supervised pretraining and supervised task learning is performed in sequential manner.}
    \label{fig:intro_compare}
    \vspace{-1.5em}
\end{figure}

This study uses functional knowledge transfer between self-supervised representation learning and other supervised downstream tasks.
Figure~\ref{fig:intro_compare} compares both knowledge transfer approaches. 
The proposed method jointly optimizes contrastive self-supervised learning with classification task learning on ResNet-50~\cite{resnet} backbone, explored on three public datasets of different visual domains, CIFAR10~\cite{cifar}, Intel Image~\cite{intel}, and Aptos~\cite{aptos}. 
The proposed approach enhances supervised task performance on all three datasets, supporting the hypothesis.
Quantitative and qualitative comparisons are made between the proposed and conventional knowledge transfer approach.
The following are the main contributions of this work:
\begin{enumerate}
    \item Explored functional knowledge transfer with self-supervised representation learning towards making it applicable to the small-batch size and small-scale dataset.
    \item Hypothesizes that self-supervised learning reinforces supervised task learning and vice versa.
\end{enumerate}
With these contributions, proposed approach improves supervised task performance on all three datasets, supported by qualitative results and provide preliminary empirical support for the hypothesis. 

\section{Related Work}
Joint embedding architecture and method based self-supervised learning has shown significant advances in label-free representation learning paradigm. It is based on learning similarity in transformed views of input images and the way it learns robust features by avoiding collapsed representation it is divided into several categories, e.g., i)~Contrastive Methods (SimCLR~\cite{simclr}, MoCo~\cite{moco}), ii)~Distillation (BYOL~\cite{byol}, SimSiam~\cite{simsiam}), iii)~Clustering (SwAV~\cite{swae}), and (iv)~Information Maximization (Barlow Twins~\cite{bt}, VICReg~\cite{vicreg}). All these methods have explored the representational knowledge transfer approach, where pretraining is performed, and learned parameters are transferred as knowledge to enable downstream tasks. However, functional knowledge transfer and simultaneous training are unexplored. Although some work has been carried out to exploit the label details in self-supervised methods~\cite{supcon}, especially contrastive learning.

On the other side, multi-task learning~\cite{mtl98, mtl_bene, mtl_survey, metamtl} has explored functional knowledge transfer by simultaneous training procedures is their natural requirement and has shown progress toward improved performance and computational efficiency.
Self-supervised learning approaches for functional knowledge transfer are unexplored.
It could make self-supervised algorithms computationally efficient and adaptable to small datasets by integrating with other learning tasks.

\section{Method}\label{sec:method}
The proposed method enables a specific type of inductive transfer, called functional knowledge transfer~\cite{inductive} on self-supervised representation learning approach by incorporating simultaneous training with downstream task learning. Specifically, the proposed method employs the contrastive learning method~\cite{simclr} for self-supervised representation learning and classification as downstream tasks on multiple datasets, CIFAR10~\cite{cifar}, Aptos~\cite{aptos}, and Intel Image~\cite{intel}. The following section describes the method in detail.

Data $D:(X,Y)$ is set of input sample pair of $(x,y)$ where $x \in \mathbb{R}^d$, is the input image data of $d$ dimensions and $y$ is corresponding human-annotation from annotation space $\mathcal{C}$. The data is defined as $D : \{(x_1, y_1),...(x_n, y_n)\} \subseteq \mathbb{R}^d \times\mathcal{C}$.

\subsection{Contrastive Self-supervised Learning}
To define the joint embedding architecture and method based self-supervised learning objective, followed in contrastive learning (SimCLR~\cite{simclr}), a set of $K$ non-leanrable transformations $\mathcal{T} : \{t_k\}_{k \in K}$ is defined, which are image processing based augmentations, provides transformed views of input image $(x\;', x\;'')$, to retain the invariant feature learning. 
Further, learnable function $f:\mathbb{R}^d$ $\rightarrow$ $\mathbb{R}^{m}$ parameterized by learnable parameters $\Theta_f$ which is Convolutional Neural Network (CNN) backbone and another learnable function $g:\mathbb{R}^m\rightarrow$ $\mathbb{R}^{\Tilde{m}}$ parameterized by learnable parameters $\Theta_g$ which is projector network is defined. With that, Noise contrastive estimation~\cite{nce} based self-supervised contrastive learning objective, NT-Xent (Normalized Temperature Scaled Cross Entropy) loss is defined in Eq.~\ref{eq:ssl_contrastive_loss}.
\begin{align} \label{eq:ssl_contrastive_loss}
    \mathcal{L}_{SSL}=\sum_{(x\;',x\;'') \in \mathcal{T}(X)} - \log{\dfrac{\mathrm{e}^{\mathcal{A}}}{{\sum_{k =1}^{2|X|} 1_{[k \neq x\;']}}\mathrm{e}^{\mathcal{B}}}}
\end{align}
\begin{equation} \label{eq:A}
    \mathcal{A}= (sim(g(\Theta_g;f(\Theta_f;x\;')), g(\Theta_g;f(\Theta_f;x\;'')))/\tau)
\end{equation}
\begin{equation} \label{eq:B}
    \mathcal{B}= (sim(g(\Theta_g;f(\Theta_f;x\;')), g(\Theta_g;f(\Theta_f;x^k)))/\tau)
\end{equation}
where, $\mathcal{A}$ defines similarity for positive pairs, $\mathcal{B}$ constitute similarity for negative pairs with denominator part of Eq.~\ref{eq:ssl_contrastive_loss}, and $sim$ is cosine similarity and $\tau$ is temperature scale parameter, and $\mathcal{L}_{SSL}$ is contrastive loss.
\subsection{Supervised Task Learning}
Supervised learning objective for mentioned downstream task of classification can be mentioned in terms of cross entropy loss $\mathcal{L}_{CE}$, defined in Eq.~\ref{eq:ce_loss}. 
\begin{equation} \label{eq:ce_loss}
    \mathcal{L}_{CE}=- \frac{1}{|D|}\sum_{(x,y) \in D} \sum_{c \in \mathcal{C}} y_{c}\log(f(\Theta;x_{c}))
\end{equation}
\subsection{Representational Knowledge Transfer}
Representational Knowledge Transfer is extensively explored in self-supervised learning, not only in contrastive learning but also in other self-supervised paradigms, i.e., distillation~\cite{simsiam, byol} and information maximization~\cite{bt,vicreg}. This type of knowledge transfer comprises two stages;
\begin{enumerate}
    \item First stage is self-supervised pretraining of CNN backbone without requiring labels which learns invariant representations of underlying visual concepts by similarity learning, described in Eq.~\ref{eq:ssl_contrastive_loss}
    \item Second stage is downstream supervised tasks learning in which learnt representations from stage one is used by initializing the learning parameters of CNN encoder, and supervised training is performed accordance to the task, e.g., classification, described in Eq.~\ref{eq:ce_loss} 
\end{enumerate}
The first part of Figure~\ref{fig:intro_compare} symbolically depicts the process.
\subsection{Functional Knowledge Transfer}
Functional Knowledge Transfer in the context of self-supervised learning is defined by jointly optimizing the self-supervised learning objective with supervised task learning objective. $\mathcal{L}_{FKT}$ loss described in Eq.~\ref{eq:fkt_loss} is single stage process where parameters learning is simultaneous and influenced by both loss objectives in just-in-time manner. $\lambda$ is parameter for balancing losses, however kept $1$ in all experiments.  
\begin{equation} \label{eq:fkt_loss}
    \mathcal{L}_{FKT}=\mathcal{L}_{CE} + \lambda~\mathcal{L}_{SSL}
\end{equation}
The second part of Figure~\ref{fig:intro_compare} symbolically demonstrate the process.
 \begin{figure*}[t] 
    \includegraphics[width=2\columnwidth, height=0.25\linewidth]{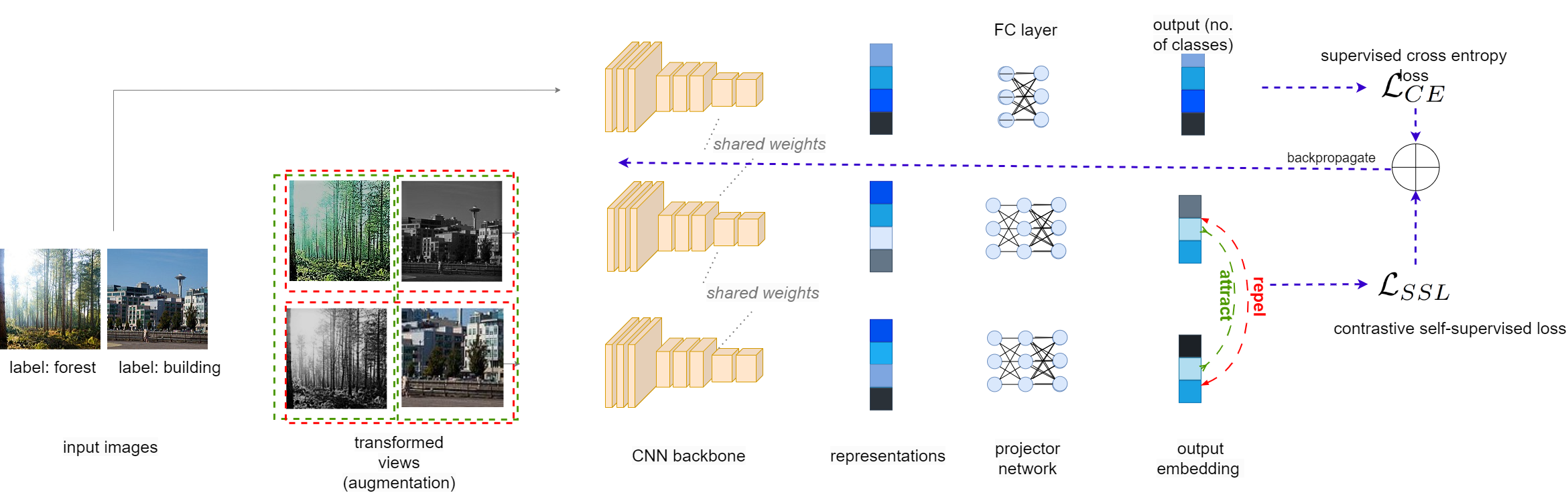}
    \centering
    \caption{Illustrates the Functional Knowledge Transfer where contrastive loss and cross entropy loss is computed on self-supervised and supervised tasks respectively and jointly backpropagated, which enables simultaneous training.}
    \label{fig:fkt}
\end{figure*}
 
\begin{figure}[t] 
    \includegraphics[width=\columnwidth]{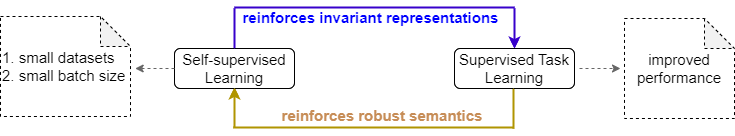}
    \centering
    \caption{Demonstrate the bi-directional constructive reinforcements for self-supervised learning and supervised task learning which enables self-supervision on relatively smaller batch size and small-scale datasets and improves classification performance.}
    \label{fig:reinforce}
\end{figure}

 \noindent
 \textit{Analytical Reasoning}: Functional Knowledge Transfer in context of self-supervised learning with supervised task learning is based on reinforced effects of tasks to each other. More concretely, it is shown in Figure~\ref{fig:reinforce} and defined as:
 \begin{itemize}
     \item Invariant Features - Self-supervised learning objective shares invariant generalized descriminative features, which reinforces the task specific feature learning for given human annotation
      \item Robust Semantics - Supervised task learning shares robust semantic information (e.g., categorization, clusters of similar concepts) of underlying visual concepts of image backed by human knowledge, which reinforces similarity learning of semantically similar visual concepts
 \end{itemize}
 This bi-directional constructive reinforcements improves learning of both the tasks, which can enable to learn contrastive learning on relatively smaller batch sizes and smaller datasets and improved performance for supervised downstream task, shown in Figure~\ref{fig:reinforce}.
 
\section{Datasets}
This study uses public datasets of natural geographic scenes, atomic objects, and medical images to investigate functional knowledge transfer on self-supervised representation learning in diverse visual concepts.
The Table~\ref{tab:dataset} summarizes the three datasets used in this work.
\begin{table}[h]
\centering
\caption{Datasets}
\label{tab:dataset}
\begin{tabular}{c|c|cl|c}
\hline
\textbf{Dataest}               & \textbf{Image type}            & \multicolumn{2}{c|}{\textbf{No. of images}}         & \textbf{No. of}                      \\ \cline{3-4}
\multicolumn{1}{l|}{\textbf{}} & \multicolumn{1}{l|}{\textbf{}} & \multicolumn{1}{l|}{\textbf{Train}} & \textbf{Test} & \multicolumn{1}{l}{\textbf{classes}} \\ \hline
CIFAR-10~\cite{cifar}          & singular objects               & \multicolumn{1}{c|}{50000}          & 10000         & 10                                   \\
Intel Image~\cite{intel}       & natural scenes                 & \multicolumn{1}{c|}{14034}          & 3000          & 6                                    \\
APTOS 2019~\cite{aptos}        & retinal images                 & \multicolumn{1}{c|}{3263}           & 399           & 5         \\ \hline                         
\end{tabular}%
\vspace{-1.5em}
\end{table}

\section{Experimental Details}
To evaluate the applicability of contrastive self-supervised learning method in functional knowledge transfer approach, detailed experimentation was performed on three public datasets, CIFAR10~\cite{cifar}, Intel Image~\cite{intel}, and Aptos~\cite{aptos} from diverse visual domains. 
Functional knowledge transfer is employed by joint training of self-supervised (simCLR~\cite{simclr}) and supervised task learning (classification) as mentioned in the Section~\ref{sec:method}. 
A comparative study is performed by bench-marking the proposed approach to conventional approach of representational knowledge transfer, where the model is pretrained and then trained for the downstream task.  

Methodological investigations are preferred; hence, common hyperparameters are configured for all three datasets with both transfer knowledge approaches.
To emphasize a less compute-intensive approach, single GPU implementation is preferred with ResNet-50~\cite{resnet} backbone and batch size of 256 for contrastive learning, which is much smaller than the original work. 
Due to this very reason, contrastive pretraining on CIFAR is perfomed with batch size 256, which was not available elsewhere.  
Pretraining, downstream task, and joint training are configured for 100 epochs. 
Self-supervised pretraining in both approaches uses LARS optimizer with learning rate 0.001 temperature scale 0.5 and employs standard augmentations suggested in the original work simCLR~\cite{simclr}. 
Supervised learning classification tasks in both approaches use SGD optimizer with a learning rate 0.025. 
All the experiments are repeated three times and the mean value of the performance metric is reported with standard deviation.
\section{Results and Discussions}
Table II describes the multi-class classification performance of the proposed approach by comparing it with the conventional approach for all three datasets. 
A consistent improvement, up to $1.40\%$, is observed in accuracy for all three datasets for the proposed functional knowledge transfer approach. 
It is worth noting that all the results show negligible standard deviation across several trial. 
The proposed approach has improved performance over previous work on APTOS and intel image datasets, also supported by qualitative analysis. 
Important observations are briefly described as follows -
\begin{table}[]
\label{tab:rn50}
\caption{Results for representational transfer (sequentially SSL pretrained then downstream task) and functional transfer (Joint optimization of self supervised pretrained and dowsntream task) on three datasets. A common architecture ResNet-50 is employed, and SimCLR contrastive pretraining with batch size of 256 only is used. \$ - \textit{produced pretraining on batch size 256 referring original work simCLR~\cite{simclr}}.}
\centering
\begin{tabular}
{@{\hskip4pt}c@{\hskip4pt}|@{\hskip4pt}c@{\hskip4pt}|@{\hskip4pt}c@{\hskip4pt}|@{\hskip4pt}c@{\hskip4pt}|@{\hskip4pt}c@{\hskip4pt}}
\hline
Dataset                       & Method                                                               & Accuracy            & Precision           & Recall              \\ \hline
\multirow{2}{*}{CIFAR10}      & \begin{tabular}[c]{@{}c@{}}Representational \\ Transfer \textsuperscript{\$} \end{tabular} & 92.20±0.11          & 92.18±0.10          & 92.21±0.10          \\
                              & \begin{tabular}[c]{@{}c@{}}Functional \\ Transfer\end{tabular}       & \textbf{93.60±0.10} & \textbf{93.62±0.13} & \textbf{93.59±0.11} \\ \hline
\multirow{2}{*}{Intel Image} & \begin{tabular}[c]{@{}c@{}}Representational \\ Transfer\end{tabular} & 93.18±0.15          & 93.15±0.18          & 93.17±0.20          \\
                              & \begin{tabular}[c]{@{}c@{}}Functional \\ Transfer\end{tabular}       & \textbf{93.70±0.13} & \textbf{93.33±0.11} & \textbf{93.31±0.11} \\ \hline
\multirow{2}{*}{Aptos 2019}   & \begin{tabular}[c]{@{}c@{}}Representational \\ Transfer\end{tabular} & 83.10±0.10          & 83.05±0.09          & \textbf{83.05±0.12} \\
                              & \begin{tabular}[c]{@{}c@{}}Functional \\ Transfer\end{tabular}       & \textbf{83.32±0.11} & \textbf{83.14±0.10} & 83.04±0.10          \\ \hline
\end{tabular}
\end{table}
\begin{table}[h]
\label{tab:rn18}
\caption{Ablation Results for Representational Transfer and Functional Transfer on intel image dataset on ResNet-18 backbone. Batch size is 256.}
\begin{tabular}{c|c|c|c|c}
\hline
Dataset                                                                  & Method                                                               & Accuracy   & Precision  & Recall     \\ \hline
\multirow{2}{*}{\begin{tabular}[c]{@{}c@{}}Intel \\ Image\end{tabular}} & \begin{tabular}[c]{@{}c@{}}Representational \\ Transfer\end{tabular} & 78.28±0.14 & 78.16±0.15 & 78.19±0.18 \\ \cline{2-5} 
                                                                         & \begin{tabular}[c]{@{}c@{}}Functional \\ Transfer\end{tabular}       & 78.52±0.13 & 78.38±0.14 & 78.43±0.12 \\ \hline
\end{tabular}
\vspace{-1.5em}
\end{table}
\begin{figure}[h]
\centerline{
\includegraphics[width=8cm, height=4cm]{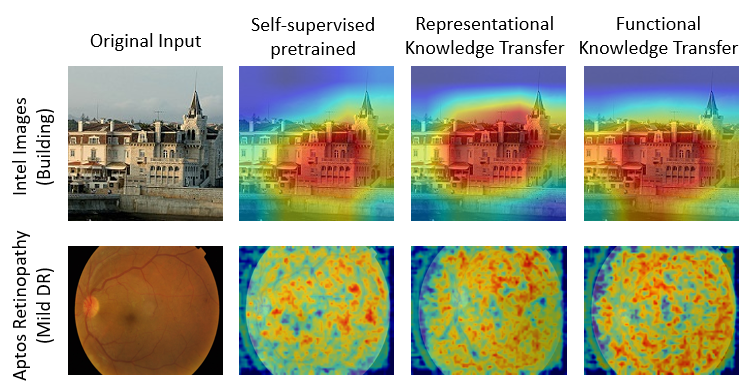} 
}
\caption{Pretrained model, representational knowledge transfer, and functional transfer approaches are compared for class activation maps (CAM). First instance is from building category from intel image dataset, and second instance is mild DR category from APTOS dataset. CAM not produced for CIFAR10 dataset due to very small size of input.
}
\label{fig:cam}
\vspace{-1.5em}
\end{figure}

\begin{figure}[!h]
\centerline{
\includegraphics[scale=0.7]{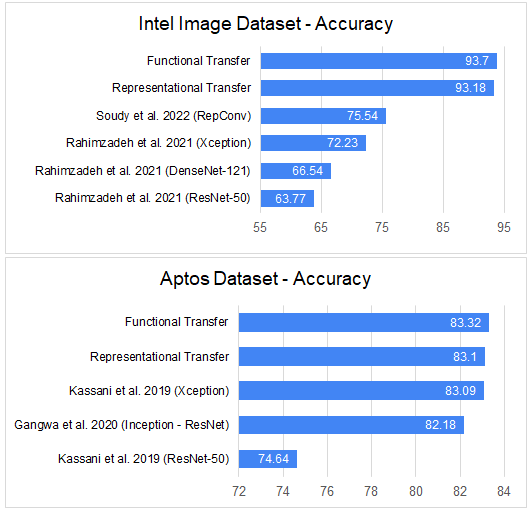} 
}
\caption{Comparison with previous works: Intel Image (top), Aptos retinopathy fundus (bottom).
}
\label{fig:sota_aptos_Intel}
\vspace{-1.5em}
\end{figure}

\textbf{Functional Knowledge Transfer improves performance}: Results comparisons in Table II clearly show the inspiring trend that functional transfer has improved the downstream task performance regardless of the dataset over the conventional approach. 
It also outperforms previous works, using the same ResNet-50 architecture and beyond, as shown in Figure~\ref{fig:sota_aptos_Intel} for APTOS and Intel Image datasets.

\textbf{Functional Knowledge Transfer enables efficient self-supervision}: Enabling self-supervised learning on small-scale datasets and smaller batch size is another significant outcome for the functional knowledge transfer approach. 
It supports the hypothesis mentioned in Figure ~\ref{fig:reinforce} where both tasks reinforced the efficiency to each other. 
However, more investigation is required in efficiently fusing self-supervised and supervised learning task loss objectives for even further improved performance.

\textbf{Functional Knowledge Transfer demonstrates computational efficiency}: Representational knowledge transfer requires 100 epochs of pretraining followed by 100 epochs of downstream supervised task learning. In contrast, the functional knowledge transfer approach performs better by joint training for 100 epochs. Effectively, functional knowledge transfer requires roughly half of computations or at-least saves downstream task computation costs. It is also essential to evaluate the functional knowledge transfer for domain adaptation and other transfer learning scenarios in future work because self-supervised learning in representation knowledge transfer intends to do the transfer learning. 

\textbf{Qualitative Robustness}: Quantitative results and performance are also supported by qualitative analysis shown through class activation maps in Figure ~\ref{fig:cam} for two datasets, Intel Image and Aptos, where attention regions are displayed. 
It clearly shows the ability to attend to the region of interest to capture the essence of visual concepts in the images.
When compared to the pretrained and representational knowledge transfer approaches, functional knowledge transfer demonstrated very competitive and even more focused attention region.

\textbf{Ablation}: Ablation study is performed on ResNet-18 backbone on Intel Image dataset (Table III), which shows marginal improvement, which gives motivation to investigate further in this direction.

\section{Conclusion}
Functional knowledge transfer is explored on contrastive self-supervised learning with classification tasks where exciting performance improvement is depicted across multiple public datasets. It has shown preliminary empirical support for enabling contrastive self-supervised learning on small batches and small-scale datasets by reinforcing the task during joint training. This study strongly encourages further investigation of functional knowledge transfer using different self-supervised learning paradigms and supervised learning tasks.

\bibliography{refs}
\bibliographystyle{IEEEtran}

\end{document}